\documentclass[10pt,twocolumn,letterpaper]{article}

\usepackage{iccv}
\usepackage{times}
\usepackage{epsfig}
\usepackage{graphicx}
\usepackage{amsmath}
\usepackage{amssymb}

\usepackage{algorithm}
\usepackage[noend]{algpseudocode}

\newlength\myfigsize

\usepackage[breaklinks=true,bookmarks=false]{hyperref}
\hypersetup{colorlinks=true,urlcolor=blue,citecolor=red,pagebackref=true,backref=page}%

\iccvfinalcopy 

\def\iccvPaperID{12665} 

\ificcvfinal\fi

\begin{document}

\title{On the De-duplication of LAION-2B}


\author{Ryan Webster\\
Unicaen\\
{\tt\small ryan.webster@unicaen.fr}
\and
Loic Simon\\
ENSICAEN\\
{\tt\small loic.simon@ensicaen.fr}
\and
Julien Rabin\\
Unicaen\\
{\tt\small julien.rabin@unicaen.fr}
\and
Frederic Jurie\\
Unicaen\\
{\tt\small frederic.jurie@unicaen.fr}}

\maketitle
\ificcvfinal\thispagestyle{empty}\fi

\begin{abstract}
   Generative models, such as DALL-E, Midjourney, and Stable Diffusion, have societal implications that extend beyond the field of computer science. These models require large image databases like LAION-2B, which contain two billion images. At this scale, manual inspection is difficult and automated analysis is challenging. In addition, recent studies show that duplicated images pose copyright problems for models trained on LAION2B, which hinders its usability. This paper proposes an algorithmic chain that runs with modest compute, that compresses CLIP features to enable efficient duplicate detection, even for vast image volumes. Our approach demonstrates that roughly 700 million, or about a third of LAION-2B's images, are duplicates. Our method also provides the histograms of duplication on this dataset, which we use to reveal more examples of verbatim copies by Stable Diffusion and further justify the approach. The current version of the de-duplicated set can be found at \url{github.com/ryanwebster90/snip-dedup}.
\end{abstract}

\section{Introduction}

The image generation models from textual descriptions, such as DALL-E \cite{ramesh2022hierarchical}, Midjourney \cite{midjourney}, or Stable Diffusion \cite{rombach2022high}, have set a significant milestone in the field of artificial intelligence, with a societal impact that extends far beyond this domain.

Recent breakthroughs in computer vision are owe much to the availability of massive image databases. These databases continue to demonstrate impressive gains in performance, even with billions of data points \cite{radford2021learning,jia2021scaling}. In fact, there are now publicly available datasets of this scale, such as LAION, that have been compiled for download \cite{schuhmann2021laion,schuhmann2022laion,kakaobrain2022coyo-700m}. These datasets have been widely used by the open-source community to develop powerful generative text-to-image models, including the popular Stable Diffusion model \cite{rombach2022high}. LAION-5B \cite{schuhmann2022laion}, in particular, marks a new era for public datasets, containing billions of image-caption pairs that have been refined by the CLIP network to ensure they are relevant to their captions. For those working mainly with English captions, LAION-2B-en is a useful subset of LAION-5B.

The availability of massive image databases has been a key factor in recent advancements in computer vision.  The qualities of these databases effects the models trained on them. Unfortunately, it has been shown that the popular Stable Diffusion text-to-image system can directly copy training images \cite{carlini2023extracting}, raising concerns about copyright infringement. Other demonstrations have shown more subtle forms of dataset copying, such as through image regions or patches \cite{somepalli2022diffusion}. It is worth noting that these issues have emerged alongside the rise of publicly available, billion-scale datasets, which were collected via highly automated web scrapers \cite{schuhmann2021laion,schuhmann2022laion}. While duplicate images can cause problems \cite{ramesh2022hierarchical}, finding duplicates or similar images within such vast datasets must be done with retrieval systems, hopefully compatible with CLIP features, as they are already computed during dataset construction. Popular tools to create clip retrieval are clip retrieval, faiss and autofaiss, employed by LAION \cite{clipretrieval,yu-2022-autofaiss,johnson2019billion}.

In this work, we study the multi-modal retrieval problem on LAION-2B and provide the following contributions:
\begin{itemize}
    \item In Sec.~\ref{sec:feat_comp}, we present a contrastive feature compression technique, which we dub Subset Nearest Neighbor CLIP compression (SNIP). We show in Sec.~\ref{sec:result_search} that this loss better retains feature semantics for multimodal tasks whilst having competitive retrieval performance.
    \item In Sec.~\ref{sec:results_dedup}, we use several of our compact indices to de-duplicate LAION-2B. We show that LAION-2B has roughly 700 million duplicated images with 91\% precision. The de-duplicated set, histograms and SNIP code can be found online at \url{github.com/ryanwebster90/snip-dedup}.
    \item Finally, in Sec.~\ref{sec:results_verb}, we show that by synthesizing only the most duplicated images, we find additional images verbatim copied by Stable Diffusion with significantly less resources than previous demonstrations.
\end{itemize}


\section{Related work}


\paragraph{The CLIP network}
The CLIP network has achieved state-of-the-art (SOTA) performance on various zero-shot and transfer tasks, as reported in \cite{radford2021learning}. This method employs a contrastive loss to align image and text feature representations, resulting in feature spaces that can be utilized in various applications. One such application is the conditioning of text-to-image models, which has been explored in several studies, including \cite{rombach2022high,xu2022versatile,sauer2023stylegan}. The open source repository, OpenCLIP~\cite{OpenClip}, has successfully reproduced the results of the original CLIP paper, albeit with significant computational resources. In fact, OpenCLIP has released several models that have surpassed the zero-shot ImageNet performance. 
\paragraph{Copyright in Generative Models}
Carlini \etal \cite{carlini2023extracting} demonstrated that Stable Diffusion can produce images that are exact copies of training images. As noted by the original authors of \cite{ramesh2022hierarchical}, duplicated images can cause such phenomena. Carlini \etal used CLIP features to filter duplicated images and detect duplicates through overly repetitive synthesis. Similarly, in \cite{somepalli2022diffusion}, it was shown that image regions can also be overfit in a similar way. In our work, we have identified several images in \cite{carlini2023extracting} that are verbatim copies and highly duplicated, and other images that are highly duplicated but not overfit or only loosely overfit.

\setlength{\myfigsize}{0.4\textwidth}
\begin{figure*}[!htb]
  \centering
    \includegraphics[width=\myfigsize]{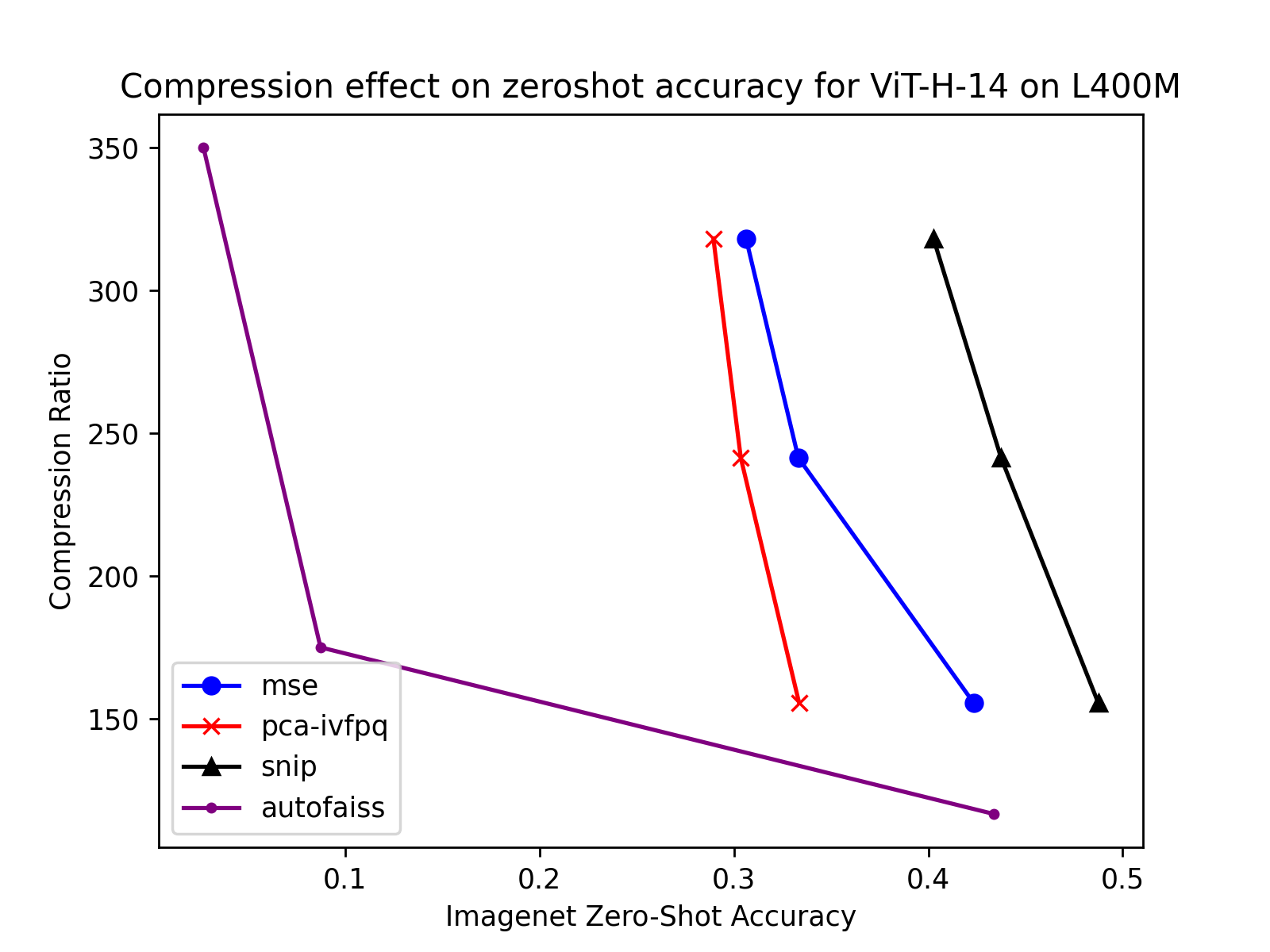}
    \includegraphics[width=\myfigsize]{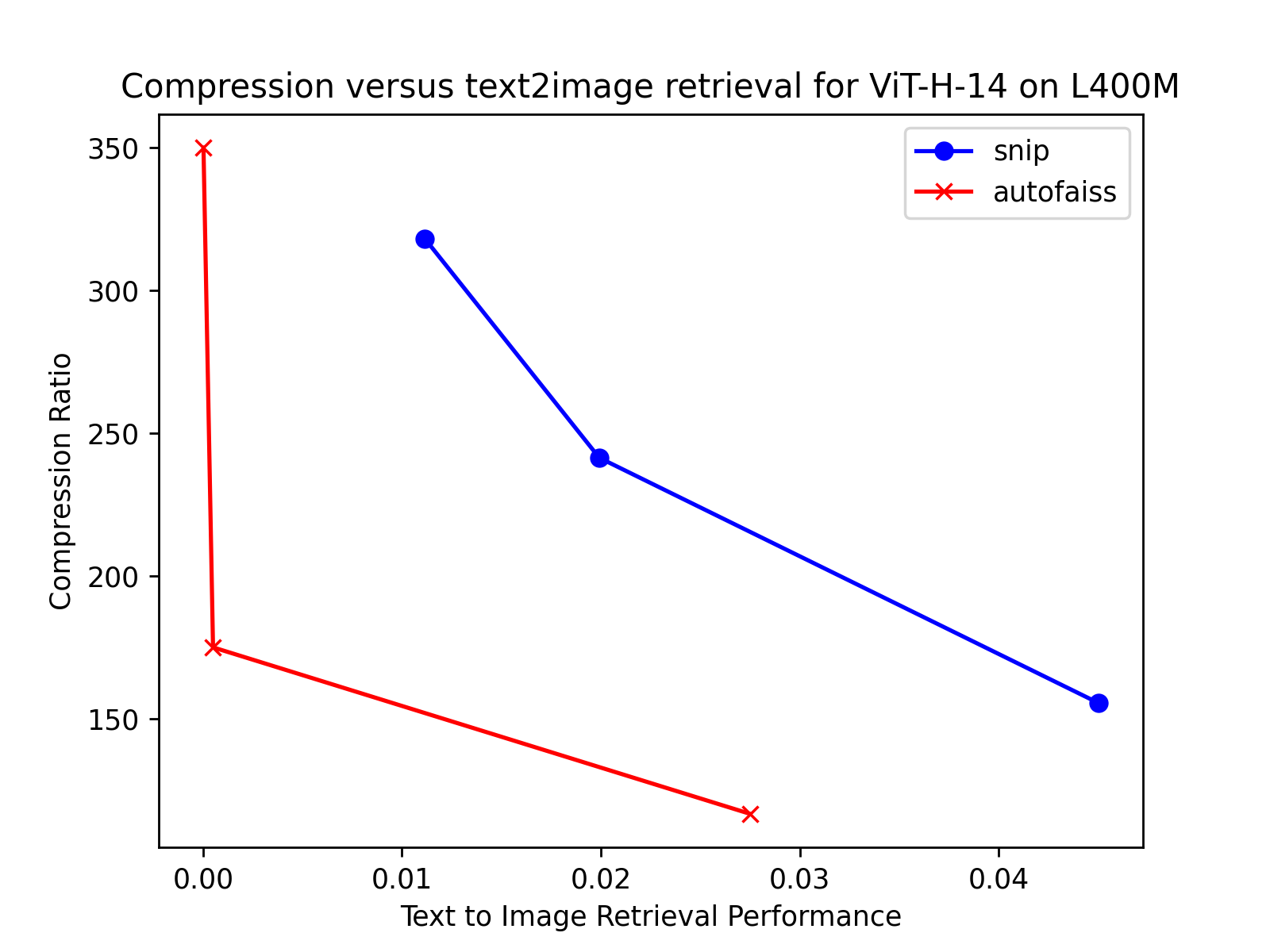}
  \caption{Investigating the effects of compression on several tasks requiring multi-modality of image features. MSE based compression (pca, mse) lose semantics for ImageNet classes at higher compression ratios, w.r.t. to the snip based losses.}
  \label{fig:multimodal failure}
\end{figure*}

\paragraph{De-duplication}
Many methods have been proposed for image de-duplication (\eg, \cite{li2021CEDedupCosteffectiveConvolutional, li2021QHashEfficientHashing}), including those that use perceptual hashes \cite{jafari2021survey} or end-to-end representations \cite{pizzi2022self}. Zhang \etal \cite{zhang2023DatasetdrivenUnsupervisedObject} have also proposed learning dataset-specific search descriptors with contrastive loss. However, these methods are not well-suited to generating compact descriptors for efficiently cleaning massive image databases such as LAION-2B (L2B), as developing solutions that require training or fine-tuning on datasets of this size presents its own set of challenges. Aware of this issue, LAION has released a set of CLIP features and a nearest neighbor index for the L2B dataset, which can be accessed through a popular web demo \cite{clipretrieval}. Although these indices are useful with enough server resources, queries would be in 50-100ms range when the index is memory mapped and thus be too slow for billions scale de-duplication. The indices adopt the widely used approximate nearest neighbor tool Faiss~\cite{johnson2019billion}, particularly the product quantization routines discussed later in this paper. Finally, we note that billions scale de-duplication is not a technical achievement per-say, but as we explore an ‘‘extreme'' compression scenario, we envision our method being potentially more accessible to the community for dataset exploration.

\section{CLIP Feature Compression}\label{sec:feat_comp}
We begin with a generic auto-encoding baseline using mean squared error, which is a standard technique for feature compression \cite{theis2017LossyImageCompression}. Denoting $(x^T,x^I)$ as minibatches of CLIP text and image features we have 
\begin{equation}
    \mathcal{L}_{\textsc{MSE}}(x^I;E^I,D^I) = \lVert x^I - D^I(E^I(x^I)) \rVert^2
    .\label{eq:mse loss}
\end{equation}
where, $D^I,E^I$ are the encoder/decoder networks for the image features. The compression rate is thus controlled by the output dimension of $E^I$, or the latent space dimension, which will be used for index creation and search later on.  

The compression can be done on either image CLIP or text CLIP descriptors. However, the limitation is that if the compression of the two modalities is done independently of each other, there is a risk of losing the alignment between the modalities. Our experiments with hybrid encoders (fusing modalities or mixing contrastive and reconstruction losses) showed worse results than this baseline or the methods presented next.

\paragraph{Contrastive Compression}  We propose a second compression method designed to preserve text and image feature alignment. To achieve this, we suggest using the original contrastive loss function proposed in \cite{radford2021learning}. There are several approaches to compressing features this way, including the use of an autoencoder or applying the clip loss directly in the original feature space. However, we found that using a "latent" clip loss as $\mathcal{L}_{\textsc{CLIP}}(E^T(x^T), E^I(x^I))$, with $E^I,E^T$ as the image and text encoders, was better overall and use this loss for all experiments.

Although using CLIP loss alone was effective in some tasks, such as zero-shot ImageNet classification, we observed that to enable more accurate nearest neighbor search, it was better to also introduce a term to maintain distance properties between neighboring elements in the dataset. Therefore, we propose a new approach that involves computing the nearest neighbors (w.r.t. image features) in a "chunk" (here, we used sliding chunks $k=10$M samples), defined as follows

\begin{equation}
\begin{array}{ll}
    \mathcal{L}_{\textsc{SNIP}}(x^T_{k},x^I_{k};E^T,E^I) = \\ 
    & \hspace*{-23mm} \mathcal{L}_\textsc{CLIP}(E^T(x^T_{k}), E^I(x^I_{k}))\\ 
    &\hspace*{-28mm} + \lambda \ \mathcal{L}_{\textsc{CLIP}}(E^I(x^I_{k}),E^I(x^I_{1\textsc{-NN}(x^I_{k}),k}) )
\end{array}
\label{eq:snip}
\end{equation}%
where $1\textsc{-NN}(x^I_{k})$ is the nearest neighbor of sample $x^I_{k}$ in the chunk $k$.   
The nearest neighbor search within each chunk is done exhaustively by measuring the L2 distances between all the pairs of the chunk.

\paragraph{Approximate Search}
Even compressed, searching through CLIP descriptors at billions scale is still too expensive.  We therefore use techniques for approximate nearest neighbor search akin to those found in Johnson \etal \cite{johnson2019billion}. 
One of the simplest and oldest techniques for fast nearest neighbor search relies on an inverted file system (IVF). First one computes the k-means centroids of the database vectors and then groups vectors to their closest centroid to form the inverted files. During search, queries only look through the $\tau$ closest inverted files by exhaustive search over the closest centroids, which is tractable because the number of centroids is typically small. We explore building a two-level quantizer on top of our compressed features, which compresses database vectors as follows:
\begin{equation}
y \approx q_{1}(y) + q_{2}(y - q_{1}(y))
\end{equation}
where $y$ is first approximated with $q_{1}$, and then its residual quantized by $q_{2}$. $k$-NN's are retrieved as those minimizing a quantity known as the asymmetric distance:
\begin{equation}
    d_{ADC}(x,y) =  \lVert x - q(y)\rVert
\end{equation}\label{eq:ivfadc}%
where $q(y)$ is a quantized database vector. In the case when $q_{1}$ is an IVF, this quantity is only minimized over database vectors mapping to the same centroid as the query or top $\tau$ centroids (referred to as the nprobe parameter). Specifically, we explore Product Quantization (PQ) as our secondary quantizer \cite{jegou2010product}. PQ splits vectors into $M$ sub-vectors, $y = [q^{0}(y^{0}),...q^{M}(y^{M}) ]$ where $M$ is an even divisor of $y's$ dimension and the sub-quantizers are again k-means. Higher values of $M$ result in higher compression ratios.

\setlength{\myfigsize}{0.32\textwidth}
\begin{figure*}[tb]
\centering
\includegraphics[width=\myfigsize]{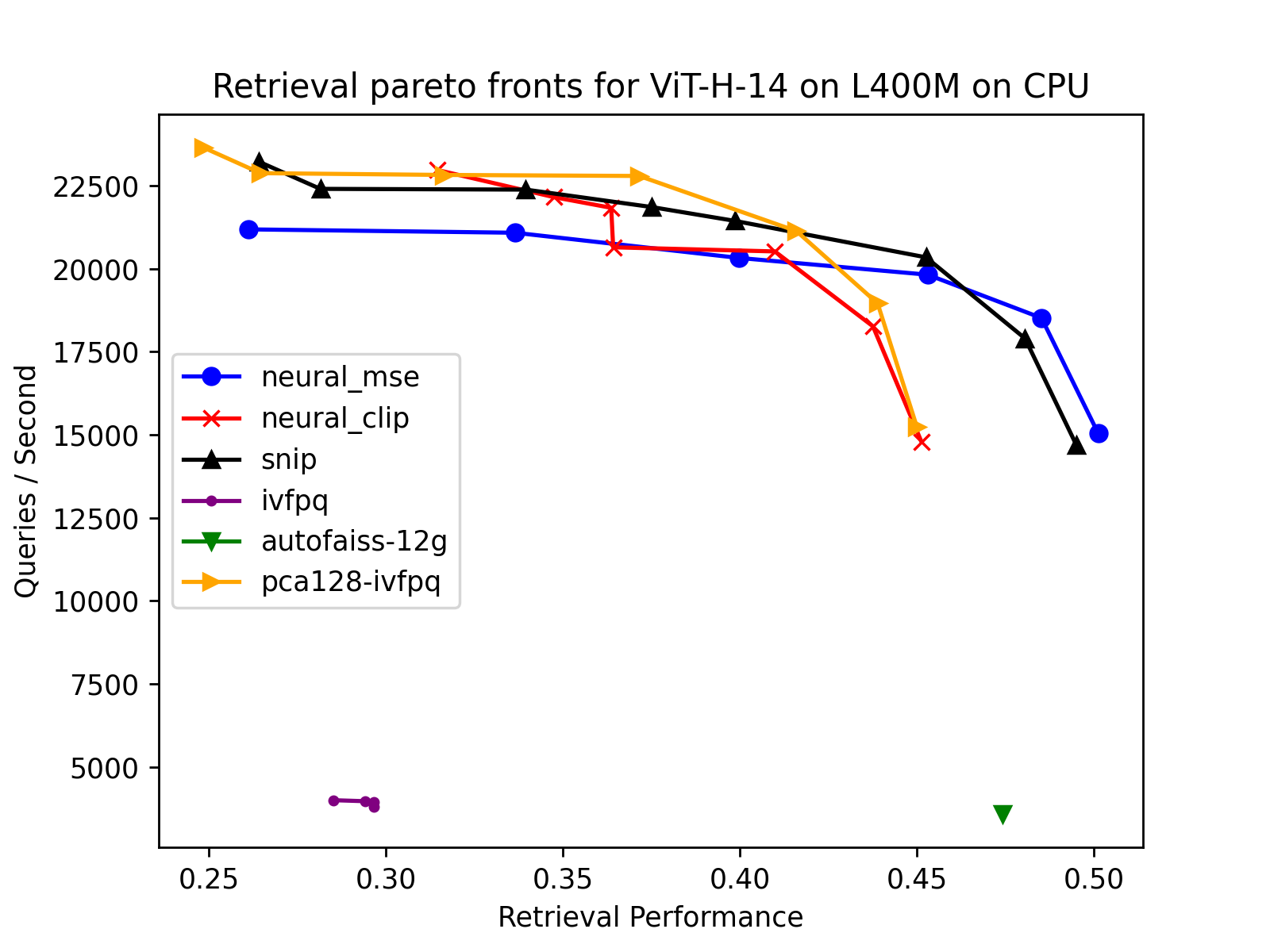}
\includegraphics[width=\myfigsize]{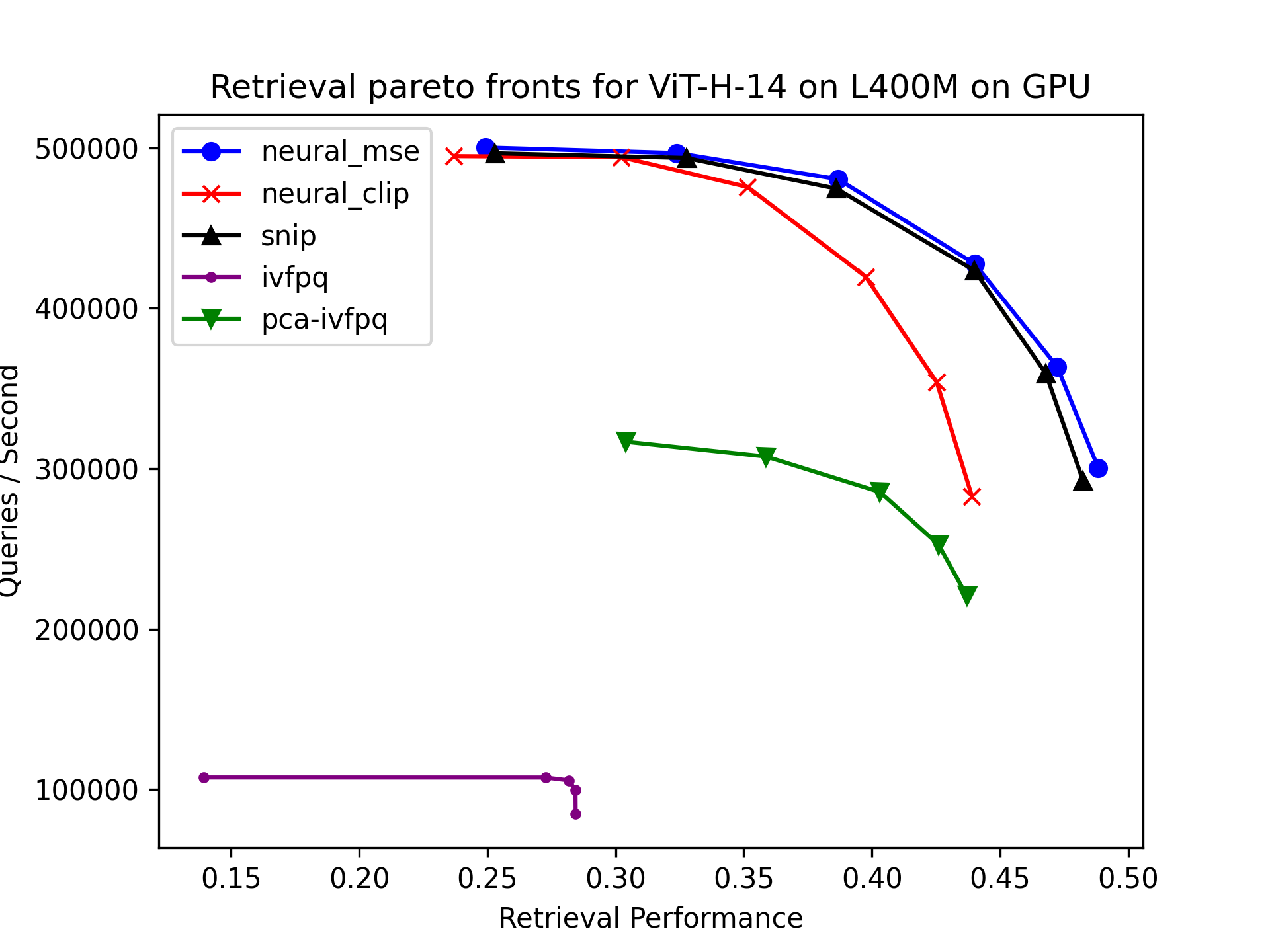}
\includegraphics[width=\myfigsize]{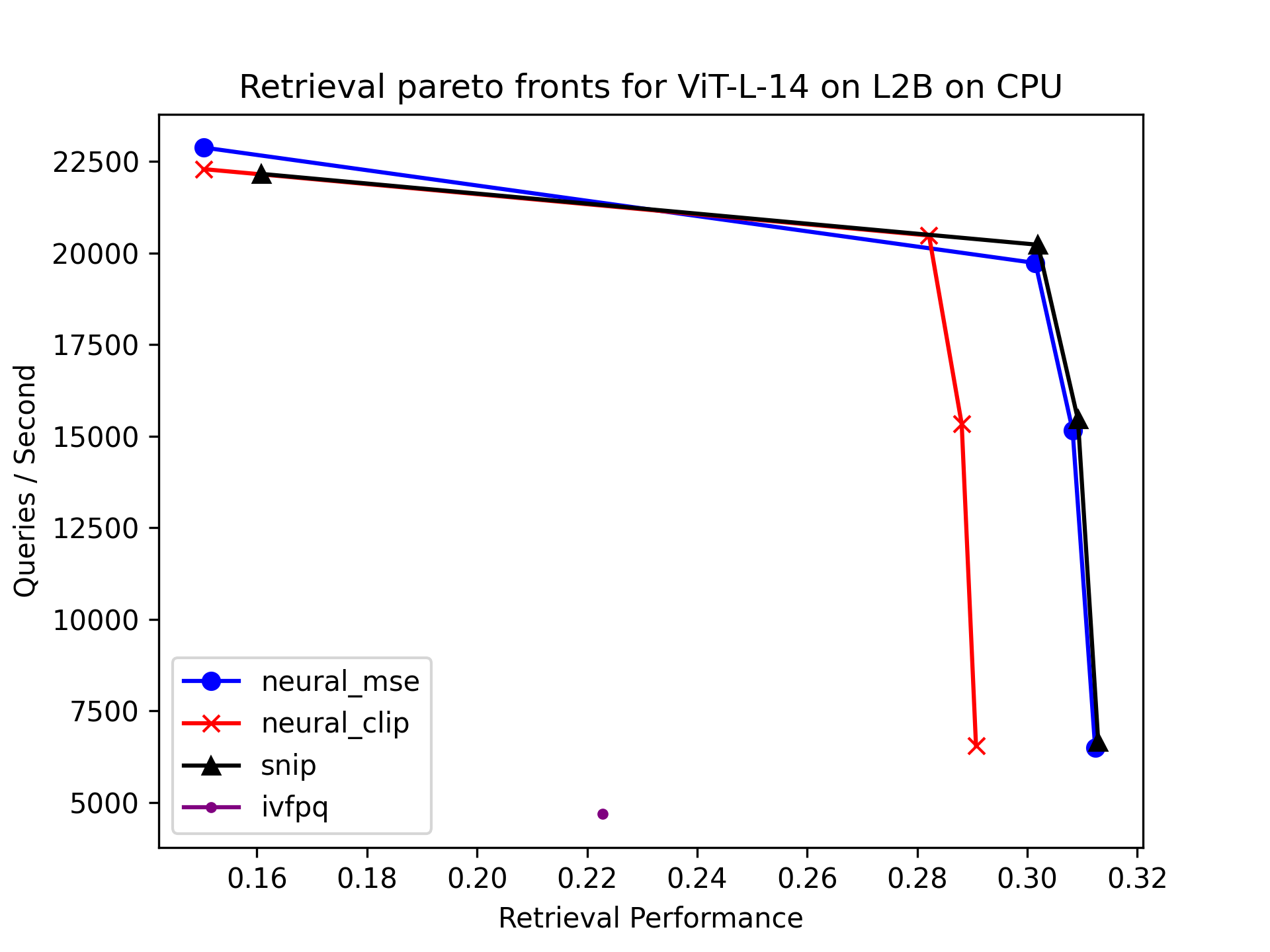}
\caption{Image feature only retrieval pareto fronts for ViT-H-14 indices on L400M on the CPU (left), GPU (middle) and L2B on CPU (right). MSE based losses perform similarly on this task compared to the contrastive ones.}
\label{fig:retrieval bench}
\end{figure*}

\section{Image Similarity Search}\label{sec:result_search}
We generate a synthetic ground truth by computing a small set of $k$-nearest neighbors using brute force. We then compared the index retrieval results with this ground truth as follows:
\begin{equation}
R \cap k := \frac{1}{N} \sum_{q \in Q} \frac{ |k\textsc{-NN}(q_{i}) \cap k\textsc{-ANN}(q_{i})| }{k} \label{eq:eval r int k}\tag{$R \cap k$}
\end{equation}
The approximate $k$-nearest neighbors ($k\textsc{-ANN}$) were computed for the index under review, while the exhaustive search on raw features ($k\textsc{-NN}$) served as the ground truth.  It's worth noting that the approach used to construct the ground truth for the (non-multimodal) Deep1B dataset \cite{babenko2016efficient} is similar to the one described in \cite{johnson2019billion}.

We did ablations over the latent space dimensions and found that encoding to a relatively high dimension of 128 first (from 1024 for ViT-H-14) was more effective, than for example encoding to a low dimension directly and giving more memory to the quantizer. We note that the approach in \cite{johnson2019billion} also used a similar dimension for the compressed space before quantization. Also like in \cite{johnson2019billion}, we used the IVFPQ index for every index we built atop of our descriptors and finally did several comparisons to a vanilla IVFPQ on raw features. 

For the IVF, we chose $2^{16}$ centroids, and for the PQ we constructed collections of indices with different values of $M$ (the number of dimension chunks) with $2^{8}$ centroids each. Finally, we highlight the popular open source tool AutoFaiss, which will automatically construct and efficient PQ index given a memory budget. All of our AutoFaiss indices use multiple PQ steps, with an initial optimized product quantization (OPQ) step for dimensionality reduction (see \cite{ge2013optimized}), as well as a hierarchical navigable world search (HSNW) \cite{malkov2018efficient}. For instance, when providing the tool with ``6G" total memory for the index, it produced a faiss construction pipeline of ``OPQ16\_112,IVF65536\_HNSW32,PQ16x8" (Please see the faiss library for more details\cite{johnson2019billion}). 
We noted these indices were not more effective on the GPU and excluded them from these benchmarks. To form Pareto fronts, we merely built a collection of indices as before, but varying the memory parameter.

\paragraph{Multi-modal Search Results}
The original CLIP network demonstrated impressive performance on the challenging zero-shot ImageNet classification task. With CLIP, this can be achieved by getting an encoding for each category $c$ with the caption ‘‘this is an image of [$c$]’’, and then performing an exhaustive search with CLIP encoded ImageNet samples via taking a maximum inner product (as is maximized for positive pairs during contrastive training). We'd like to not only explore how well our descriptors preserve multi-modal information, but the quantized representation in the index as well. For this, we reconstruct database vectors using the \texttt{sa\_encode} and \texttt{sa\_decode} functions of the faiss library, and then do multi-modal search on reconstructed vectors.
We compare to AutoFaiss (see previous section), which is used for multi-modal search hosted by LAION \cite{schuhmann2022laion}, but in a more extreme compression regime.
 
Figure~\ref{fig:multimodal failure} shows the performance of each index type versus the "compression ratio," which is simply the uncompressed size of all database features divided by the index size (plus a negligible size of the encoding network). Again, we create a collection of indices with varying $M$ parameters to form the IVFPQ pareto fronts. The $\mathcal{L}_{\textsc{MSE}}$ descriptor indices zero-shot performance are clearly less efficient than the $\mathcal{L}_{\textsc{SNIP}}$ ones. Autofaiss performs the worst at these compression ratios (in general we saw a sharp drop in the AutoFaiss performance above a certain compression ratio for every task). For text-to-image search, our snip model also performs better than AutoFaiss. 
We now move on to a wider variety of comparisons. 

\paragraph{Image to Image Search Results}
Figure~\ref{fig:retrieval bench} presents the retrieval benchmarks for the L400M and L2B datasets. We assessed each method across a range of benchmarks and created Pareto fronts by varying the  number of chunks for product quantization (i.e., $M=2,4,8$) and the n probe parameter ($\tau=1,2,4$). Notably, MSE and SNIP performed similarly on image similarity search, while CLIP-only loss networks consistently performed worse. However, as mentioned, the CLIP networks (both CLIP and SNIP) do perform better on multi-modal tasks. Thus, SNIP can be seen as a best of both worlds descriptor for these types of indices. Figure~\ref{fig:retrieval bench} also indicates that indices similar to those in \cite{Johnson2021BillionScaleSS}, using a PCA as a first compression and IVFPQ as a quantizer, have some drop off in performance, as with the CLIP only indices.

\section{De-duplicating LAION 2B}\label{sec:results_dedup}

In this section, we investigate several of the index creation pipelines discussed previously to de-duplicate LAION-2B. Our focus on more "extreme" compression up until now is justified; de-duplicating such a large database quickly relies on a fast and thus compact index. The goal of this effort was to merely find many duplicates with decent precision quickly, rather than having a perfect de-duplication.
\par
Our indices return the asymmetric distance~\eqref{eq:ivfadc}, which can result in non-zero distances even for in-database queries. Additionally, since the secondary quantizer operates globally over the IVF, different centroids may contain significantly different $d_{ADC}$ distributions. Consequently, we found that applying an absolute threshold on $d_{ADC}$ alone was insufficient for effective de-duplication performance.

Instead of using an absolute threshold on the $d_{ADC}$, which was found to be insufficient for good de-duplication performance due to the global nature of the secondary quantizer over the IVF, an adaptive threshold based on the $d_{ADC}$ between a query vector and its quantized version was used. This approach takes into account that duplicated images may not necessarily have a low absolute $d_{ADC}$, but they will all have the same $d_{ADC}$. Specifically, duplicates $y\in k\textsc{-NN}(x)$ of a query image $x$ are identified if 
\begin{equation}\label{eq:threshold_NN}
    \frac{\lvert d_{ADC}(x,x) - d_{ADC}(x,y) \rvert}{d_{ADC}(x,x)} < T_{ADC}.
    \tag{Dup}
\end{equation} 
It is worth noting that this threshold is applied only retrieved $k$-NN's. The duplication detection performance was consistent with the previous retrieval experiments, and the best performing indices were selected to perform the de-duplication of LAION2B.

Table~\ref{tab:dedup} presents the results of de-duplication experiments conducted on the LAION2B dataset using the ViT-L-14 and ViT-H-14 CLIP feature sets. To estimate the precision, we randomly selected 100k labeled duplicated images and computed a ground truth with their raw CLIP features. We chose a conservative threshold on the raw features, which would still only label near duplicates (via inspection). De-duplication was performed on a machine with 32 GB of RAM and took only several days. Finally, we also estimate precision and recall of merely indicating the presence of one duplicate (See Table~\ref{fig:dedup_pareto}) via exhaustive search, and again comparing to Eq.~\ref{eq:threshold_NN}. 

\begin{figure}[ht]
\centering
\includegraphics[width=.9\linewidth]{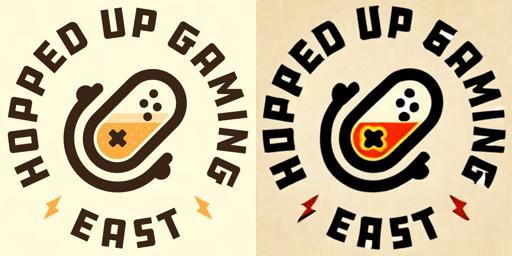}
\includegraphics[width=.9\linewidth]{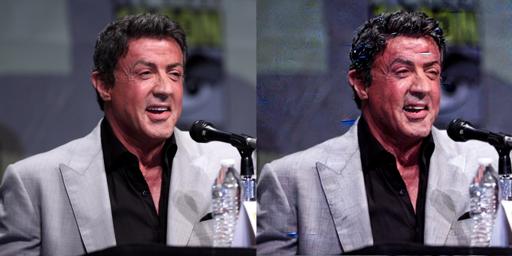}
\caption{We find several new verbatim copied images by Stable Diffusion (SD), by synthesizing the most copied images on L2B. Left are images duplicated database images and right are synthesized by SD. Top image is highly duplicated (more than 5k times), bottom image is moderately duplicated (2000 times).} 
\label{fig:verbatim}
\end{figure}

\setlength{\myfigsize}{0.6\textwidth}
\begin{figure}[tb]
\centering
\includegraphics[width=\myfigsize]{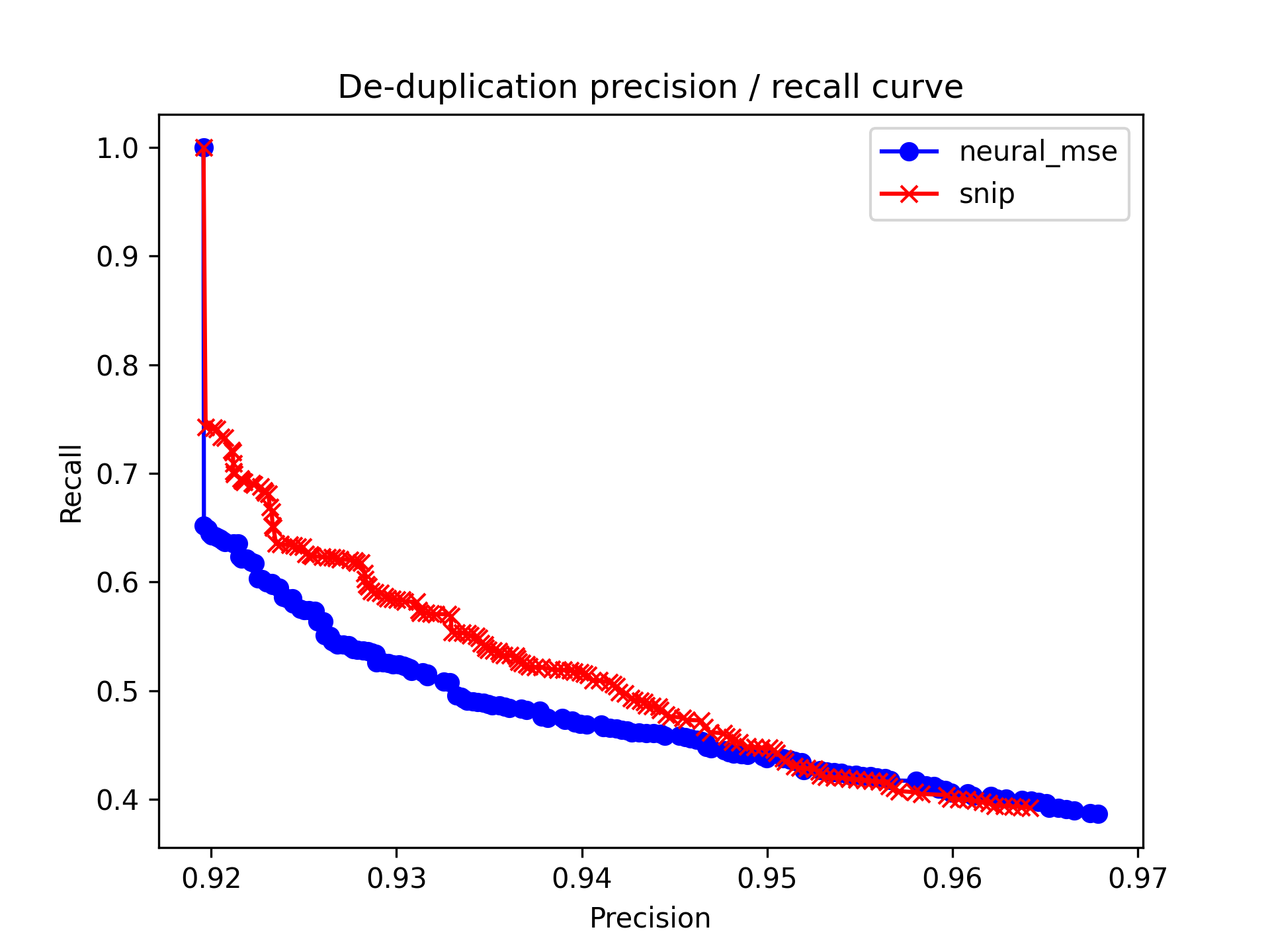}
\caption{Here, we compute an exhaustive nearest neighbor search for 10k samples. Then we use the raw clip features as a duplication ground truth. Out of 10k samples, 9k were marked as near duplicates by the raw clip features (ViT-L-14). The above figure demonstrates that after a certain threshold, everything is marked as a duplicate. The precision is high merely because the labels are highly imbalanced. This shows that our method should miss some duplicates over the whole set. In practice, we choose an adaptive threshold of 1\%.}
\label{fig:dedup_pareto}
\end{figure}
\begin{table}[tb]
\centering
    \begin{tabular}{|c|c|c|}
    \hline
    Index & Duplicates Found & Precision  \\
    \hline
    ViT-L-14 MSE (GPU,M16) &  624,636,782 & .91 \\
    \hline
    ViT-L-14 SNIP (GPU,M16) &  640,402,551 & .92 \\
    \hline
    ViT-H-14 SNIP (CPU,M4) &  682,146,751 & .91 \\
    \hline
    \end{tabular}
    \vspace{1mm}
    \caption{De-duplication of L2B. The top two indices were sharded over 4-GPUs, with $M$=16 and the ViT-L-14 network. Below is on ViT-H-14 features, on the cpu and not sharded. Note, the precisions aren't directly comparable here as we use different raw clip thresholds for each. Image space comparisons will be indicated in future versions so they are comparable.}
    \smallskip
\end{table}\label{tab:dedup}

To actually compute the de-duplication, we loop over the entire dataset using algorithm \ref{alg:representative_list}, that can be used with any index. We note this algorithm will also yield the representative sets, that can be used to group duplicated images together.
\begin{algorithm}
\caption{Deduplication of a dataset $\mathcal D$}\label{alg:representative_list}
\begin{algorithmic}
\Require index for $k$-NN search 
\State Set $k$ the number of retrieved nearest neighbors
\State \text{Initialize list of representatives by self reference:} $\forall\ i\;  R_i \leftarrow i$ 
\For{$x_i \in \mathcal D$}
    \State $\mathcal S \leftarrow k\text{-NN}(x_i)$ 
    \State $\mathcal S' \leftarrow \{ x_j \in \mathcal S | j>i \text{ and } x_j \text{ satisfies } \eqref{eq:threshold_NN} \}$
\For{$x_j \in \mathcal S'$}
    \State $R_j \leftarrow R_i$
\EndFor
\EndFor
\end{algorithmic}
\end{algorithm}

\section{Investigating Verbatim Copies in Stable Diffusion}\label{sec:results_verb}

Recently, \cite{carlini2023extracting} demonstrated stable diffusion can verbatim copy training images on certain prompts. Similar to \cite{carlini2023extracting}, our objective is to generate exact replicas of images in the training set using the stable diffusion generator with their corresponding prompts. However, the approach in  \cite{carlini2023extracting} is rather brute forcedl; they used duplication as only a weak filter to select candidates for copied images and synthesized nearly 175 million images. They yielded just 94 images; finding images is thus incredibly rare and akin to finding a needle in a hay stack. Not having the same computational resources, we concentrate on a subset of one hundred images from the most duplicated images on L2B, i.e. selecting iamges from the long tail of the duplication histogram.

We utilize the prompt corresponding to each chosen image as input for , generating a total of 9 random samples. Upon reviewing several hundred generated samples manually, we discovered numerous images that are verbatim copies of training data  
(as illustrated in Fig.~\ref{fig:verbatim}). For the Sylvester Stallone image (with the strange prompt ``Rambo 5 und Rocky Spin-Off - Sylvester Stallone gibt Updates"), we found stable diffusion will completely ignore the seed, and only generate this image. In addition, we identified several images that were highly duplicated but not copied verbatim, as well as others that were nearly identical and fell somewhere in between, as previously observed in \cite{somepalli2022diffusion}.
A promising avenue for future research involves identifying which images are more susceptible to replication than others. The number of duplicates in the database appears to be just one factor among several. We speculate that the variability of prompts linked to the duplicates, as well as the distinctiveness of certain prompts, could also play a role.

\section{Conclusions}
In this work, we discussed an algorithmic chain which has de-duplicated LAION-2B-en with modest resources and decent precision. We release code to download this de-duplicated datasets, their representative sets (See Alg.~\ref{alg:representative_list}, the duplication histograms and finally SNIP indices for dataset exploration. This is important for dataset usability; recent work demonstrates generative models have serious copyright issues when duplicates are present. Our work uses a novel pipeline which does not require pre-training or re-computation of billions of clip features and should provide an accessible tool to the community for greater dataset transparency.

\bibliographystyle{ieee_fullname}
\bibliography{bib}

\begin{thebibliography}{10}\itemsep=-1pt

\bibitem{midjourney}
\url{https://www.midjourney.com/}.

\bibitem{babenko2016efficient}
Artem Babenko and Victor Lempitsky.
\newblock Efficient indexing of billion-scale datasets of deep descriptors.
\newblock In {\em Proceedings of the IEEE Conference on Computer Vision and
  Pattern Recognition}, pages 2055--2063, 2016.

\bibitem{clipretrieval}
Romain Beaumont.
\newblock Clip retrieval.
\newblock \url{https://github.com/rom1504/clip-retrieval}, 2022.

\bibitem{kakaobrain2022coyo-700m}
Minwoo Byeon, Beomhee Park, Haecheon Kim, Sungjun Lee, Woonhyuk Baek, and
  Saehoon Kim.
\newblock Coyo-700m: Image-text pair dataset.
\newblock \url{https://github.com/kakaobrain/coyo-dataset}, 2022.

\bibitem{carlini2023extracting}
Nicholas Carlini, Jamie Hayes, Milad Nasr, Matthew Jagielski, Vikash Sehwag,
  Florian Tram{\`e}r, Borja Balle, Daphne Ippolito, and Eric Wallace.
\newblock Extracting training data from diffusion models.
\newblock {\em arXiv preprint arXiv:2301.13188}, 2023.

\bibitem{ge2013optimized}
Tiezheng Ge, Kaiming He, Qifa Ke, and Jian Sun.
\newblock Optimized product quantization.
\newblock {\em IEEE transactions on pattern analysis and machine intelligence},
  36(4):744--755, 2013.

\bibitem{OpenClip}
Gabriel Ilharco, Mitchell Wortsman, Ross Wightman, Cade Gordon, Nicholas
  Carlini, Rohan Taori, Achal Dave, Vaishaal Shankar, Hongseok Namkoong, John
  Miller, Hannaneh Hajishirzi, Ali Farhadi, and Ludwig Schmidt.
\newblock Openclip repository.
\newblock July 2021.

\bibitem{jafari2021survey}
Omid Jafari, Preeti Maurya, Parth Nagarkar, Khandker~Mushfiqul Islam, and
  Chidambaram Crushev.
\newblock A survey on locality sensitive hashing algorithms and their
  applications.
\newblock {\em arXiv preprint arXiv:2102.08942}, 2021.

\bibitem{jegou2010product}
Herve Jegou, Matthijs Douze, and Cordelia Schmid.
\newblock Product quantization for nearest neighbor search.
\newblock {\em IEEE transactions on pattern analysis and machine intelligence},
  33(1):117--128, 2010.

\bibitem{jia2021scaling}
Chao Jia, Yinfei Yang, Ye Xia, Yi-Ting Chen, Zarana Parekh, Hieu Pham, Quoc Le,
  Yun-Hsuan Sung, Zhen Li, and Tom Duerig.
\newblock Scaling up visual and vision-language representation learning with
  noisy text supervision.
\newblock In {\em International Conference on Machine Learning}, pages
  4904--4916. PMLR, 2021.

\bibitem{johnson2019billion}
Jeff Johnson, Matthijs Douze, and Herv{\'e} J{\'e}gou.
\newblock Billion-scale similarity search with {GPUs}.
\newblock {\em IEEE Transactions on Big Data}, 7(3):535--547, 2019.

\bibitem{Johnson2021BillionScaleSS}
Jeff Johnson, Matthijs Douze, and Herv{\'e} J{\'e}gou.
\newblock Billion-scale similarity search with gpus.
\newblock {\em IEEE Transactions on Big Data}, 7:535--547, 2021.

\bibitem{li2021CEDedupCosteffectiveConvolutional}
Xuan Li, Liqiong Chang, and Xue Liu.
\newblock {{CE-Dedup}}: {{Cost-effective}} convolutional neural nets training
  based on image deduplication.
\newblock In {\em 2021 {{IEEE}} Intl Conf on Parallel \& Distributed Processing
  with Applications, Big Data \& Cloud Computing, Sustainable Computing \&
  Communications, Social Computing \& Networking
  ({{ISPA}}/{{BDCloud}}/{{SocialCom}}/{{SustainCom}}), New York City, {{NY}},
  {{USA}}, September 30 - Oct. 3, 2021}, pages 11--18. {IEEE}, 2021.

\bibitem{li2021QHashEfficientHashing}
Xuan Li, Liqiong Chang, and Xue Liu.
\newblock {{QHash}}: {{An}} efficient hashing algorithm for low-variance image
  deduplication.
\newblock In {\em 2021 {{IEEE}} 23rd Int Conf on High Performance Computing \&
  Communications; 7th Int Conf on Data Science \& Systems; 19th Int Conf on
  Smart City; 7th Int Conf on Dependability in Sensor, Cloud \& Big Data
  Systems \& Application ({{HPCC}}/{{DSS}}/{{SmartCity}}/{{DependSys}})}, pages
  9--15, 2021.

\bibitem{malkov2018efficient}
Yu~A Malkov and Dmitry~A Yashunin.
\newblock Efficient and robust approximate nearest neighbor search using
  hierarchical navigable small world graphs.
\newblock {\em IEEE transactions on pattern analysis and machine intelligence},
  42(4):824--836, 2018.

\bibitem{yu-2022-autofaiss}
Victor Paltz, Bokai Yu, and Romain Beaumont.
\newblock Autofaiss, 2022.

\bibitem{pizzi2022self}
Ed Pizzi, Sreya~Dutta Roy, Sugosh~Nagavara Ravindra, Priya Goyal, and Matthijs
  Douze.
\newblock A self-supervised descriptor for image copy detection.
\newblock In {\em Proceedings of the IEEE/CVF Conference on Computer Vision and
  Pattern Recognition}, pages 14532--14542, 2022.

\bibitem{radford2021learning}
Alec Radford, Jong~Wook Kim, Chris Hallacy, Aditya Ramesh, Gabriel Goh,
  Sandhini Agarwal, Girish Sastry, Amanda Askell, Pamela Mishkin, Jack Clark,
  et~al.
\newblock Learning transferable visual models from natural language
  supervision.
\newblock In {\em International Conference on Machine Learning}, pages
  8748--8763. PMLR, 2021.

\bibitem{ramesh2022hierarchical}
Aditya Ramesh, Prafulla Dhariwal, Alex Nichol, Casey Chu, and Mark Chen.
\newblock Hierarchical text-conditional image generation with clip latents.
\newblock {\em arXiv preprint arXiv:2204.06125}, 2022.

\bibitem{rombach2022high}
Robin Rombach, Andreas Blattmann, Dominik Lorenz, Patrick Esser, and Bj{\"o}rn
  Ommer.
\newblock High-resolution image synthesis with latent diffusion models.
\newblock In {\em Proceedings of the IEEE/CVF Conference on Computer Vision and
  Pattern Recognition}, pages 10684--10695, 2022.

\bibitem{sauer2023stylegan}
Axel Sauer, Tero Karras, Samuli Laine, Andreas Geiger, and Timo Aila.
\newblock Stylegan-t: Unlocking the power of gans for fast large-scale
  text-to-image synthesis.
\newblock {\em arXiv preprint arXiv:2301.09515}, 2023.

\bibitem{schuhmann2022laion}
Christoph Schuhmann, Romain Beaumont, Richard Vencu, Cade Gordon, Ross
  Wightman, Mehdi Cherti, Theo Coombes, Aarush Katta, Clayton Mullis, Mitchell
  Wortsman, et~al.
\newblock Laion-5b: An open large-scale dataset for training next generation
  image-text models.
\newblock {\em arXiv preprint arXiv:2210.08402}, 2022.

\bibitem{schuhmann2021laion}
Christoph Schuhmann, Richard Vencu, Romain Beaumont, Robert Kaczmarczyk,
  Clayton Mullis, Aarush Katta, Theo Coombes, Jenia Jitsev, and Aran
  Komatsuzaki.
\newblock Laion-400m: Open dataset of clip-filtered 400 million image-text
  pairs.
\newblock {\em arXiv preprint arXiv:2111.02114}, 2021.

\bibitem{somepalli2022diffusion}
Gowthami Somepalli, Vasu Singla, Micah Goldblum, Jonas Geiping, and Tom
  Goldstein.
\newblock Diffusion art or digital forgery? investigating data replication in
  diffusion models.
\newblock {\em arXiv preprint arXiv:2212.03860}, 2022.

\bibitem{theis2017LossyImageCompression}
Lucas Theis, Wenzhe Shi, Andrew Cunningham, and Ferenc Husz{\'a}r.
\newblock Lossy image compression with compressive autoencoders.
\newblock In {\em 5th International Conference on Learning Representations,
  {{ICLR}} 2017, Toulon, France, April 24-26, 2017, Conference Track
  Proceedings}. {OpenReview.net}, 2017.

\bibitem{xu2022versatile}
Xingqian Xu, Zhangyang Wang, Eric Zhang, Kai Wang, and Humphrey Shi.
\newblock Versatile diffusion: Text, images and variations all in one diffusion
  model.
\newblock 2022.

\bibitem{zhang2023DatasetdrivenUnsupervisedObject}
Zhongyan Zhang, Lei Wang, Yang Wang, Luping Zhou, Jianjia Zhang, and Fang Chen.
\newblock Dataset-driven unsupervised object discovery for region-based
  instance image retrieval.
\newblock {\em IEEE Trans. Pattern Anal. Mach. Intell.}, 45(1):247--263, 2023.

\end{thebibliography}

\end{document}